\documentclass[letterpaper, 10 pt, journal, twoside]{IEEEConf}
\usepackage[utf8]{inputenc}
\usepackage{url}
\usepackage{cite}
\usepackage{amsmath,amssymb,amsfonts}
\usepackage{textcomp}
\usepackage{lettrine}
\usepackage{caption}
\usepackage{subcaption}
\usepackage[export]{adjustbox}
\newcommand{\subparagraph}{}
\usepackage{leftidx}
\usepackage{siunitx}
\usepackage{multirow}
\usepackage{balance}
\usepackage{booktabs}

\usepackage{hyperref}
\usepackage{algorithm2e}
\RestyleAlgo{ruled}
\usepackage{mathtools}
\usepackage{setspace}

\begin{document}

\title{\LARGE \bf Faster Optimization in S-Graphs Exploiting Hierarchy
}

\author{Hriday Bavle$^{1}$, Jose Luis Sanchez-Lopez$^{1}$, Javier Civera$^{2}$, Holger Voos$^{1}$ 
\thanks{*This work was partially funded by the Fonds National de la Recherche of Luxembourg (FNR), under the project C22/IS/17387634/DEUS, and by the Spanish Government under Grant PID2021-127685NB-I00 and by the Arag{\'o}n Government under Grant DGA T45 17R/FSE.
For the purpose of Open Access, the author has applied a CC BY public copyright license to any Author Accepted Manuscript version arising from this submission.}%
\thanks{$^{1}$Authors are with the Automation and Robotics Research Group, Interdisciplinary Centre for Security, Reliability and Trust, University of Luxembourg. Holger Voos is also associated with the Faculty of Science, Technology and Medicine, University of Luxembourg, Luxembourg.
\tt{\small{\{hriday.bavle, joseluis.sanchezlopez, holger.voos\}}@uni.lu}}%
\thanks{$^{2}$Author is with I3A, Universidad de Zaragoza, Spain
{\tt\small jcivera@unizar.es}}%
}

\maketitle

\begin{abstract}

3D scene graphs hierarchically represent the environment appropriately organizing different environmental entities in various layers. Our previous work on situational graphs \cite{s_graphs}, \cite{s-graphs+} extends the concept of 3D scene graph to SLAM by tightly coupling the robot poses with the scene graph entities, achieving state-of-the-art results. Though, one of the limitations of \textit{S-Graphs} is scalability in really large environments due to the increased graph size over time, increasing the computational complexity. 

To overcome this limitation in this work we present an initial research of an improved version of \textit{S-Graphs} exploiting the hierarchy to reduce the graph size by marginalizing redundant robot poses and their connections to the observations of the same structural entities. Firstly, we propose the generation and optimization of room-local graphs encompassing all graph entities within a room-like structure. These room-local graphs are used to compress the \textit{S-Graphs} marginalizing the redundant robot keyframes within the given room. We then perform windowed local optimization of the compressed graph at regular time-distance intervals. A global optimization of the compressed graph is performed every time a loop closure is detected. We show similar accuracy compared to the baseline while showing a 39.81\% reduction in the computation time with respect to the baseline. 
\end{abstract}

\section{Introduction}

\begin{figure}[ht]
    \centering
    \includegraphics[width=0.45\textwidth]{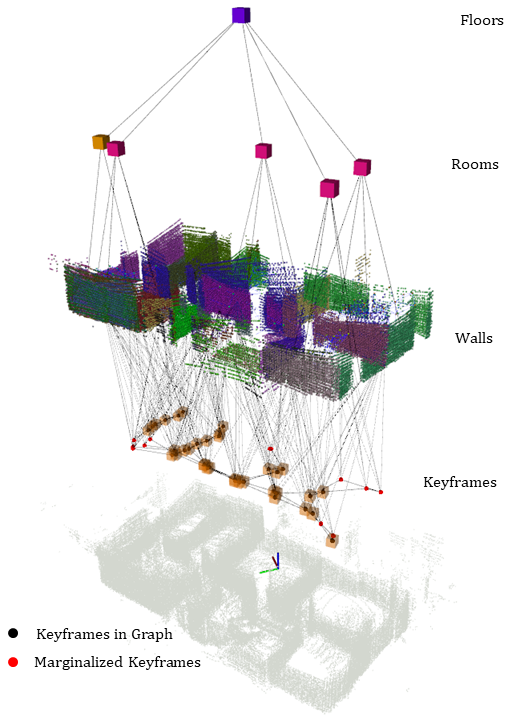}
    \caption{Generation of \textit{S-Graph} utilizing the improved optimization technique. The redundant keyframes within a given room are marginalized (red spheres) after performing room-local optimization. The remaining keyframes (black spheres) along with their connected walls-rooms-floors, undergo local and global optimization steps. The orange boxes over the keyframes highlight the \textit{S-Graph} undergoing the global optimization step where all the keyframes are utilized except the marginalized ones.}  
    \label{fig:first_page}
\end{figure} 

\begin{figure*}[!ht]
    \centering
    \includegraphics[width=1\textwidth]{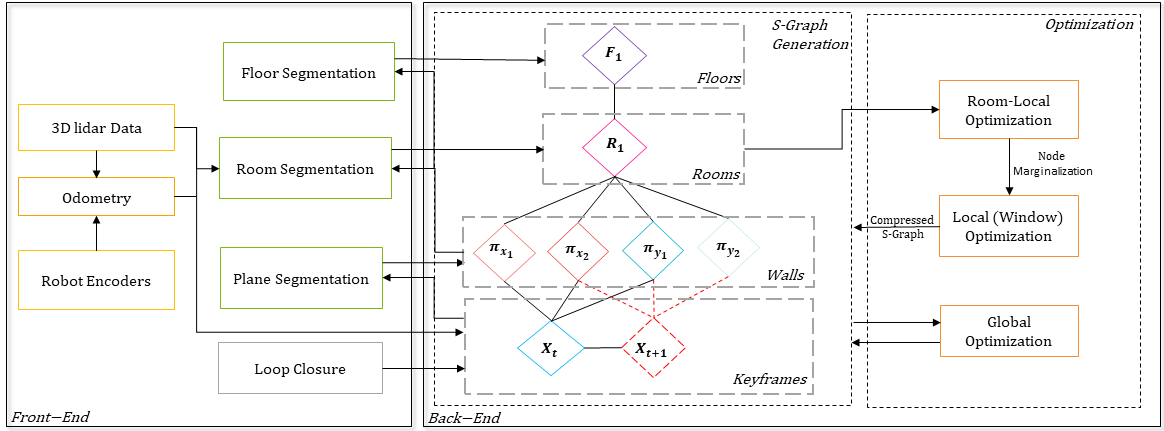}
    \caption{System architecture of the presented approach, divided into front-end and back-end. The front-end comprises of modules for generating the four-layered \textit{S-Graph}. In the back-end, the generated \textit{S-Graph} undergoes different optimization steps. In this example keyframe $X_{t+1}$ belonging to the room is marginalized out to compress the graph.}  
    \label{fig:system_architecture}
\end{figure*}

\textit{S-Graphs} \cite{s_graphs}, \cite{s-graphs+} enable mobile robots with a deep understanding of their situation through the real-time generation of a 3D scene graph tightly coupled to the robot poses in the form of a four-layered optimizable graph. \textit{S-Graphs} consists of a keyframes layer comprising the robot poses, a walls layer encompassing the wall-plane entities, a rooms layer constraining the wall-planes of appropriate rooms, and a floors layer constraining all the rooms within a given floor level. As the robot explores large-scale environments, this optimizable graph increases in the number of keyframes and semantic elements thus increasing the computational complexity and time.  

To overcome the mentioned limitation, in this preliminary work, we present an approach for exploiting the inherent hierarchy extracted by the baseline \textit{S-Graphs} to reduce the computational complexity while maintaining the robot pose and its map accuracy. Every time a room is detected all the enclosed robot keyframes, along with its attached wall-planes, and the room node form a room-local graph which is locally optimized and then utilized to marginalize the redundant keyframe nodes. In this current work, each room is kept with only one robot pose, the rest are set to marginalized. This marginalization is propagated to the entire \textit{S-Graph} to generate a compressed version of the graph. 
Furthermore, we perform optimization of the compressed graph in two steps, local and global optimization. The local optimization is performed over a window of keyframes as the robot explores the environment while the global optimization over the complete compressed graph is performed only on loop closures. Initial experimentation performed over simulated and real data show reduced computational complexity of our proposed optimization framework while maintaining the robot pose and map accuracy. Fig.~\ref{fig:first_page} highlights the presented approach where the underlying \textit{S-Graph} undergoes a global optimization step that excludes the redundant keyframes (colored in red) within a given room structure.

\begin{figure*}[!t]
\centering
\begin{subfigure}[t]{.3\textwidth}
\centering
\includegraphics[width=0.9\textwidth]{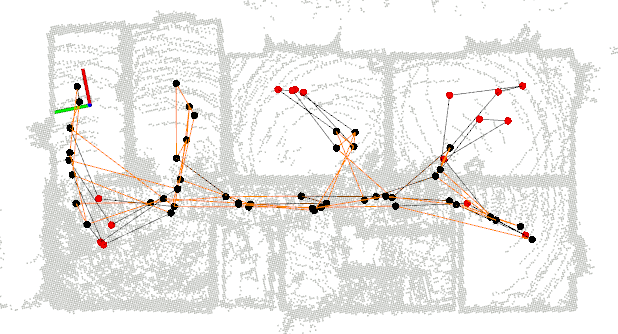}
\caption{{Proposed}}
\label{fig:s_graph_improved}
\end{subfigure}
\begin{subfigure}[t]{0.3\textwidth}
\centering
\includegraphics[width=0.9\textwidth]{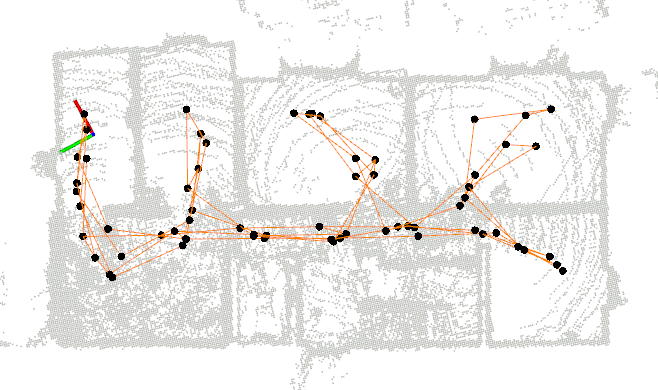}
\caption{\textit{S-Graphs+} \cite{s-graphs+}}
\label{fig:s_graphs+}
\end{subfigure}
\begin{subfigure}[t]{0.27\textwidth}
\centering
\includegraphics[width=0.9\textwidth]{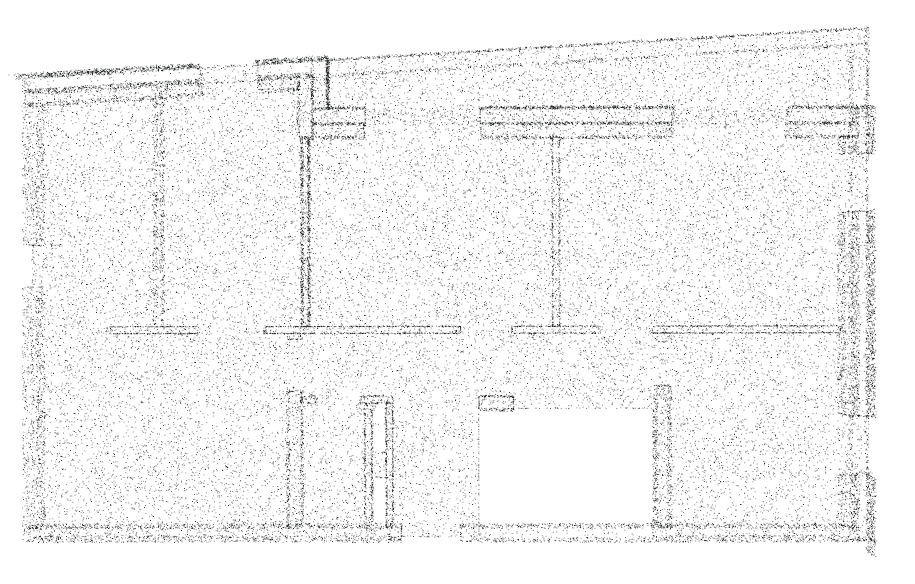}
\caption{Ground Truth}
\label{fig:s_graphs+}
\end{subfigure}
\caption{Qualitative results of a real-world experiment showing the map accuracy by our proposed work and baseline \textit{S-Graphs+} with respect to the ground truth map. The black spheres are the keyframes while the red spheres in our proposed work are the marginalized keyframes belonging to the rooms.}
\label{fig:real_c1f2}
\end{figure*}

\section{Related Work}

Several methods exist in the literature for performing factor graph compression maintaining the map accuracy while bonding the computational complexity. \cite{isam} presents an incremental soothing and mapping approach considering the information gain of measurement to the state estimates, thus instead of solving the entire graph only solving parts of the graph affected by the new measurements.  
\cite{hierarchical_optimization} presents a hierarchical optimization technique grouping nodes into different subgraphs at a lower level based on simple distance-based criteria. These subgraphs form a reduced higher-level graph which is then optimized quickly and efficiently.   
Authors \cite{information_theoretic_compression} present an information-theoretic approach for factor graph compression where laser scans measurements and their corresponding robot poses are removed such that the expected loss of information with respect to the current map is minimized.

In recent years, 3D scene graphs have emerged as an efficient way to represent the environment,  \cite{s-graphs+}, \cite{3-d_scene_graph}, \cite{hydra}. Although these works efficiently represent the environment in different hierarchical levels, there can be redundant information stored at different levels which is neither removed nor compressed, which can lead to increased computational complexity while exploring very large-scale environments. Inspired by factor graph compression techniques, in this work, we present an ongoing research work of marginalizing information from 3D scene graphs exploiting its underlying hierarchy.

\section{Proposed Approach}

Fig.~\ref{fig:system_architecture} present an overview of the proposed work. The robot odometry and 3D LiDAR information are utilized for generating the four-layered \textit{S-Graph}. The point cloud information mapped at each robot keyframe is utilized to extract wall planes, creating the walls layer. The extracted walls are utilized by the room detector to extract rooms, and by the floor detector to extract floors. During the creation of the \textit{S-Graph} the underlying optimization step can be divided into three different stages namely 1. Room-Local Optimization 2. Local Optimization 3. Global Optimization, all explained in the following sections. 

\subsection{Room-Local Optimization}

Room-Local optimization generates a local \textit{S-Graph} lying within the given structure of a room, whenever a room is detected. To generate a room-local \textit{S-Graph} we first look for a subset keyframes $K_s \subset K$ which are bounded by the walls of a four-wall room ${R}_j$ which satisfy the following condition:
\begin{gather}
     \forall \boldsymbol{n}_i \in {R}_j, \quad \boldsymbol{n_i} \cdot (\boldsymbol{k}_i - \boldsymbol{\leftidx{^M}{\Pi}}_i) < 0
\end{gather}
where $\boldsymbol{\leftidx{^M}{\Pi}}_i = \boldsymbol{n}_i \cdot d_i$ and $\boldsymbol{n}$, $d$ are the normal and the distance of wall-plane $i$. $\boldsymbol{k}_i$ is the translation of the keyframe $K_i$. Furthermore, we add in the room-local \textit{S-Graph} all the wall-planes $\boldsymbol{\pi}_i$ connected keyframe subset $K_{s}$. Lastly, we incorporate the room vertex $R_j$ and the floor $F_l$. Inspired from \cite{orbslam2}, all the keyframes outside the room but connected to the wall-planes $\boldsymbol{\pi}_i$ are incorporated in room-local graph but kept constant during optimization.   

The room-local \textit{S-Graph} is then optimized to perform a room-local optimization step which optimizes only a subset of the entire graph. Since all the robot poses within a given room observe the same structure, only the first robot pose within is room maintained. Other robot poses can be safely marginalized from the graph along with their observations. This optimized and marginalized information is used to create a compressed version of the \textit{S-Graph}.  

\subsection{Local Optimization}

The marginalized information from the room-local optimization step is included to generate a compressed \textit{S-Graph} excluding the marginalized keyframes and their edges. Marginalization leads to a generation of a disconnected graph. These disconnections are between the marginalized keyframes and their non-marginalized neighbors. To obtain a connected graph, we check for the $k$ edges $e_k$, $k=\{1, \dots, k\}$ of the $i\textsuperscript{th}$ marginalized keyframe $K_i$ with $n$ non-marginalized neighbors $K_{n}$, $n=\{1, \dots, n\}$ and connect the closest non-marginalized neighbors $K_{1}$ and $K_{2}$ with a new edge $e_f$. The information matrix of the new edge is the summation of the information matrices of the eliminated edges in $e_k$.     

After the generation of the compressed \textit{S-Graph}, the local optimization performs a windowed optimization over the last $N$ subset of the keyframes $K_N$ and their connected walls, rooms, and floors within the compressed graph. The additional keyframes not in $K_N$ that observe the walls and rooms are included in the optimization but kept constant.

\subsection{Global Optimization}

Global optimization considers the optimization of the entire compressed \textit{S-Graph} without the marginalized keyframes and their edges, if a loop closure is detected between the robot keyframes. In global optimization, only the first keyframe is fixed for optimization stability.

\section{Experimental Validation}

We validate our approach on different simulated and real datasets captured using a velodyne 3D LiDAR and a legged robotic platform. Currently, we compare our approach with the baseline \textit{S-Graphs+} \cite{s-graphs+}. 
In both the simulated and real experiments, the robot performs several random and repeated trajectories in indoor environments with different room configurations. For the simulated experiments, we report the ATE of the robot pose along with the computation time required during the optimization steps. For real experiments, we provide qualitative results in terms of map quality generated by our approach and the baseline while comparing its computation time. 

Table.~\ref{tab:simulated_results} shows the results obtained on two simulated datasets when comparing them against the baseline \textit{S-Graphs+} \cite{s-graphs+}. As can be seen from the table, our proposed method is able to provide better or similar accuracy in terms of the ATE when compared with the baseline, but given our proposed method is able to improve the computation time of the global optimization with respect to the baseline by $33\%$.       

Figure.~\ref{fig:real_c1f2} shows qualitative results of the obtained map during the execution of the real experiment. As can be appreciated from the figure our proposed approach is able to provide the same map quality when comparing it with the baseline and the ground truth map. But in this experiment as well, the average computation of \textit{S-Graphs+} is $129$ ms while our proposed approach has an average computation time of $60$ ms.
 
\begin{table}[]
\caption{ATE (cm) and the computation time [ms] of \textit{S-Graphs+} and our proposed method on simulated datasets. Best results are boldfaced.}
\centering
\small
\begin{tabular}{l | l l | l l}
\toprule
& \multicolumn{2}{c|}{\textbf{ATE} (cm) $\downarrow$} & \multicolumn{2}{c}{\parbox{1.7cm}{\textbf{Computation Time} (mean) [ms]} $\downarrow$} \\ 
\cmidrule{2-3} \cmidrule{4-5}
& \multicolumn{2}{l}{\textbf{Dataset}} \\ 
\toprule
\textbf{Method} & \textit{C1F1} & \textit{C1F2} & \textit{C1F1} & \textit{C1F2} \\ \midrule
\textit{S-Graphs+} \cite{s-graphs+} & 4.20  & \textbf{1.59} & 71.0 & 69.0 \\ 
Proposed (\textit{ours}) & \textbf{4.09} &  1.73 &  \textbf{53.0} & \textbf{41.0} \\
\bottomrule
\end{tabular}
\label{tab:simulated_results}
\end{table}

\section{Conclusion}

In this paper, we present an initial research strategy for compression and faster optimization for the real-time generation of situational graphs. We exploit the inherent hierarchical representations generated by \textit{S-Graph} to generate room-local optimizable graphs, which are used to generate a compressed version of the underlying \textit{S-Graph}. We also propose to break down the optimization of the compressed graph into two steps namely, local (windowed) optimization and global optimization. We perform preliminary experiments to compare our approach to the baseline \textit{S-Graphs+} achieving similar results in terms of accuracy but improved computation time. As for future works, we would further validate our presented approach over large-scale indoor environments with multiple floors incorporating the floor-local optimization step in the graph in addition to the room-local optimization step. 
\bibliographystyle{IEEEtran}
\bibliography{Bibliography}

\end{document}